# Dynamic Anomaly Identification in Accounting Transactions via Multi-Head Self-Attention Networks

**Yi Wang,** Columbia University, New York, USA
**Ruoyi Fang,** Golden Gate University, San Francisco, USA
**Anzhuo Xie,** Columbia University, New York, USA
**Hanrui Feng,** University of Chicago, Chicago, USA
**Jianlin Lai*,** Babson College , Wellesley, USA

*Abstract-This study addresses the problem of dynamic anomaly detection in accounting transactions and proposes a real-time detection method based on a Transformer to tackle the challenges of hidden abnormal behaviors and high timeliness requirements in complex trading environments. The approach first models accounting transaction data by representing multi-dimensional records as time-series matrices and uses embedding layers and positional encoding to achieve low-dimensional mapping of inputs. A sequence modeling structure with multi-head self-attention is then constructed to capture global dependencies and aggregate features from multiple perspectives, thereby enhancing the ability to detect abnormal patterns. The network further integrates feed-forward layers and regularization strategies to achieve deep feature representation and accurate anomaly probability estimation. To validate the effectiveness of the method, extensive experiments were conducted on a public dataset, including comparative analysis, hyperparameter sensitivity tests, environmental sensitivity tests, and data sensitivity tests. Results show that the proposed method outperforms baseline models in AUC, F1-Score, Precision, and Recall, and maintains stable performance under different environmental conditions and data perturbations. These findings confirm the applicability and advantages of the Transformer-based framework for dynamic anomaly detection in accounting transactions and provide methodological support for intelligent financial risk control and auditing.*

**CCS CONCEPTS: Computing methodologies~Machine learning~Machine learning approaches**

*Keywords: Accounting flow; anomaly detection; self-attention mechanism; risk control*

## I. INTRODUCTION

In modern economic activities, corporate accounting transactions, as a direct reflection of daily operations and capital flows, are becoming increasingly complex and frequent. With the rapid development of information technology and digitalization, large volumes of transaction records are continuously generated in real time[1]. This situation poses significant challenges to traditional audit methods that rely on manual review or rule-based detection. Manual examination and fixed-threshold approaches are not only inefficient but also show delays and instability when faced with diverse and complex abnormal behaviors. Abnormal transactions in accounting flows, whether caused by human error, system defects, or potential fraudulent actions, can seriously threaten corporate financial health and operational security. Therefore, building a system capable of efficiently and accurately identifying dynamic anomalies in real time has become an essential issue in corporate risk control and compliance management[2].

At the same time, with global economic integration and the acceleration of business activities, transaction data have become large in scale, diverse in type, and highly time-dependent. Corporate accounting flows not only include traditional income and expenditure records but also involve cross-regional, multi-currency, and multi-channel interactions. These features bring two major challenges to anomaly detection. First, the diversity and concealment of anomaly patterns make them difficult to capture quickly within massive normal transactions. Second, the growing demand for real-time processing requires enterprises to identify risks instantly when transactions occur, to prevent the accumulation of anomalies from creating systemic harm. In this context, single methods based only on statistical features are no longer sufficient. There is an urgent need to adopt more intelligent and dynamic frameworks to address these challenges[3].

From a methodological perspective, the rise of deep learning, especially sequence modeling techniques, offers new solutions for dynamic anomaly detection systems [4-7]. Accounting flows show clear time-series characteristics. Each record is related not only to current business logic but also to historical behavioral patterns [8]. Traditional recurrent neural networks can handle such time dependence to some extent, but they face bottlenecks in modeling long sequences and capturing complex dependencies [9]. Recently, models based on self-attention mechanisms have overcome these limitations by enabling parallel computation and global dependency modeling [10]. They show strong expressive power in large-scale and long-span sequence processing [11]. Applying such models to real-time dynamic anomaly detection in accounting flows can achieve global understanding and efficient aggregation of multi-dimensional features, thereby improving accuracy and robustness.

Furthermore, the study of dynamic anomaly detection in accounting flows has significance beyond safeguarding a single enterprise. It directly relates to the stable operation of the entire economic system. The presence of abnormal transactions may trigger chain reactions, leading to distorted financial reports, biased investment decisions, or even the spillover of financial risks. For small and medium enterprises, effective anomaly detection can reduce financial management costs and improve resource allocation efficiency. For large corporations and financial institutions, it enhances compliance, market trust, and competitiveness. More importantly, real-time detection enables intervention before risks spread, achieving prevention rather than post-event remediation. This forward-looking control greatly strengthens overall risk resilience.

In summary, research on real-time dynamic anomaly detection of accounting flows using Transformer models aligns with the trend of financial digitalization and intelligent governance. It responds to the urgent demand of enterprises for risk management and compliance in large-scale transaction scenarios. This direction not only advances the application of time-series modeling methods in financial data research but also provides practical tools for corporate risk prevention. Such studies can promote the integration of intelligent auditing and digital governance, and lay the foundation for building a safer, more transparent, and more efficient financial ecosystem.

## II. Related work

In existing research frameworks, anomaly detection in accounting transactions has long relied on rule-based methods. These methods typically identify abnormal transactions by setting thresholds, logical rules, or pattern matching. They are simple to implement and easy to interpret, but their limitations are clear. When facing large-scale and diverse transaction data, fixed rules struggle to adapt to dynamic environments, leading to high rates of false negatives or false positives. In real-time scenarios, such approaches often fail to capture complex anomaly patterns in time and cannot meet enterprise needs for risk prevention. Therefore, while rule-based methods had value in early studies and applications, they are inadequate in today's complex financial environment and have opened space for the development of intelligent techniques[12].

With the rise of machine learning [13-15], statistical modeling, and large language models for anomaly detection have gained attention[16-21]. These approaches learn decision boundaries between normal and abnormal cases from historical transaction data, and then use classifiers or clustering models to judge new accounting records. Their advantage lies in capturing more complex feature patterns than handcrafted rules. However, they rely heavily on labeled data[22]. In practice, abnormal samples are scarce and unevenly distributed, which makes supervised models difficult to train effectively. Unsupervised methods help mitigate the shortage of labels, but they remain vulnerable to noise and imbalanced distributions when dealing with high-dimensional and multimodal features [23]. These approaches have brought new technical perspectives to anomaly detection, yet they still show weaknesses in real-time performance and generalization[24]. In recent years, deep learning has opened new paths for anomaly detection. Convolutional neural networks and recurrent neural networks have played important roles in sequence analysis. They enhance the recognition of complex anomaly patterns through automatic feature extraction and temporal modeling [25]. However, traditional deep models struggle with long-term dependencies and often lack computational efficiency, making them unsuitable for real-time detection [26]. In addition, limited model interpretability has hindered applications in finance and accounting. Even so, deep learning has accelerated the progress of this field. It has acted as a bridge between traditional approaches and new architectures and has provided useful experience for later studies based on attention mechanisms[27].

Recent advances in dynamic anomaly detection and risk assessment have increasingly leveraged deep sequence modeling and attention-based architectures. Transformer frameworks, well-known for their capacity to model long-range temporal dependencies, have been shown to significantly improve fraud and anomaly detection in transactional scenarios by capturing complex sequential relationships in financial time series [28]. At the same time, graph neural networks have emerged as powerful tools for modeling structural dependencies within data, with entity-aware GNNs enabling automatic extraction of key associations and relations in complex environments [29]. Complementary to this, heterogeneous network learning has proven effective in discovering implicit relational patterns—such as hidden corporate ties—through multi-relational graph structures [30].

With the rapid development of graph attention mechanisms, these techniques are now widely used to refine fraud detection models by selectively aggregating important neighborhood information, thus supporting precise abnormality localization [31]. Integration of GNNs with temporal sequence learning further extends these capabilities, enabling simultaneous modeling of temporal dynamics and structural connectivity, which is particularly valuable for compliance monitoring and anomaly identification in evolving systems [32]. The fusion of Transformer and GNN components continues to gain traction, offering robust representations for unsupervised anomaly discovery in heterogeneous and high-dimensional environments [33].

Attention mechanisms themselves have become a cornerstone of modern time series modeling. By adaptively focusing on salient input signals, attention-based models can better distinguish risk signals in fluctuating or non-stationary financial data [34]. Attention-augmented recurrent networks take this further by providing improved forecasting accuracy and the ability to model both short- and long-term dependencies in financial applications [35]. Hybrid neural networks that combine LSTM, CNN, and Transformer modules allow for multiscale and multimodal learning, improving robustness and accuracy in volatility forecasting and dynamic anomaly detection [36]. Multi-head attention, initially prominent in NLP and microservice modeling, has also demonstrated its effectiveness for capturing multi-perspective dependencies in financial data streams [37].

In addition to attention and graph-based strategies, collaborative learning paradigms—such as integrating feature attention with temporal modeling—help address distributed risk assessment in multi-tenant or collaborative financial systems [38]. Temporal modeling approaches that exploit explicit structural information can provide fine-grained tracking of patient or transaction state progression, as evidenced by structure-aware networks for sequential prediction [39]. The rise of contrastive learning further enriches feature representations by encouraging the model to distinguish subtle differences between normal and abnormal patterns, which is especially valuable in unsupervised or weakly supervised detection scenarios [40-41]. As data privacy and distributed computation become more important, federated anomaly detection frameworks have gained attention for supporting personalized modeling across decentralized platforms without direct data exchange [42]. To enhance robustness, recent studies have begun to incorporate causal modeling, which improves resistance to spurious correlations by focusing on underlying cause-effect relations in fault and anomaly scenarios [43]. Reinforcement learning methods such as deep Q-learning have also been introduced for workflow optimization and adaptive audit strategies, providing data-driven decision-making in complex environments [44]. Knowledge-driven feature engineering, including function-based neural modeling, further strengthens model interpretability and domain adaptability [45].

Finally, temporal graph attention and structured information extraction are increasingly used for both clinical and financial anomaly detection, providing unified representations that link temporal, structural, and semantic dependencies [46]. Collectively, these methodological advances form the technical foundation for next-generation, real-time anomaly detection and intelligent financial risk control.

## III. METHOD

In this study, the method was designed to achieve real-time dynamic anomaly detection in accounting transactions through efficient sequence modeling and multi-dimensional feature fusion. The model architecture is shown in Figure 1.

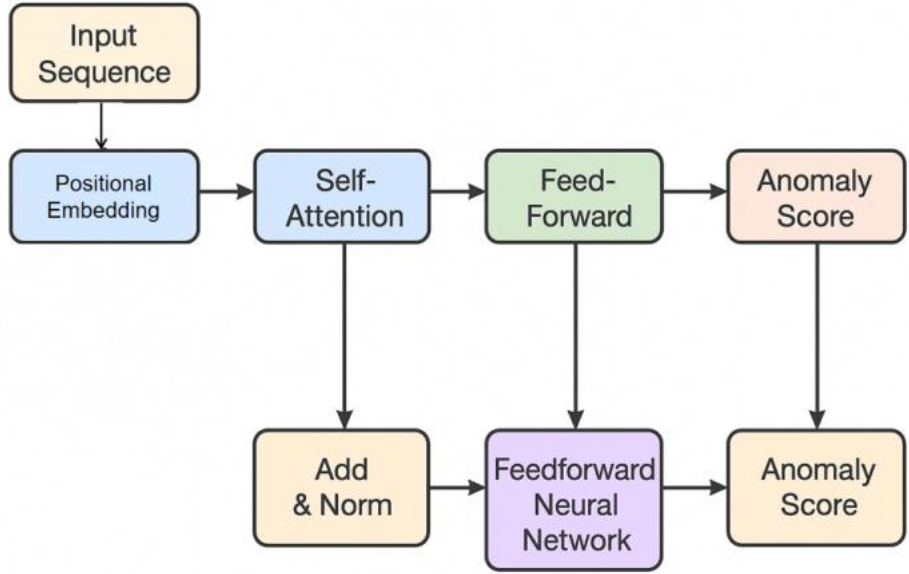

Figure 1. Overall model architecture

First, the original accounting flow data is represented as a time series matrix. Let the transaction sequence be $X = \{x_1, x_2, ..., x_T\}$, where each $x_t \in R^d$ represents a d-dimensional feature vector at time t. To ensure that the features can reflect the temporal dependency in the model input, the original input is embedded into the latent space using linear transformation and position encoding:

$$H_0 = XW_e + P \qquad (1)$$

Where $W_e \in R^{d \times d_h}$ is the feature embedding matrix, $P \in R^{T \times d_h}$ is the position encoding matrix, and $d_h$ is the latent space dimension. This step ensures that the model can capture both feature distribution and temporal order.

At the core of the modeling, we introduce a Transformer architecture based on the self-attention mechanism. Its core principle is to measure the dependencies between different time slices using an attention score matrix. For any input hidden state H, we first obtain the query, key, and value matrices through a linear mapping:

$$Q = HW_Q, \quad K = HW_K, \quad V = HW_V \qquad (2)$$

Where $W_Q, W_K, W_V$ is the learnable parameter matrix. The self-attention mechanism calculates the weights by scaling the dot product:

$$Attention(Q, K, V) = \text{softmax}(\frac{QK^T}{\sqrt{d_k}})V \qquad (3)$$

This mechanism enables the model to capture long-range dependencies between different transaction records globally, thereby identifying potential abnormal patterns.

To further enhance the aggregation of multi-dimensional features and multi-angle modeling, a multi-head attention mechanism is introduced. By dividing the latent space into h subspaces, multiple attention results are calculated in parallel, and then spliced and linearly mapped. The formula is as follows:

$$MultiHead(H) = Concat(head_1, ..., head_h)W_O \qquad (4)$$

Where each $W_O \in R^{hd_k \times d_h}$ is an output mapping matrix. This design ensures that the model can understand the characteristic patterns of transaction sequences from different perspectives, effectively alleviating the information shortage problem caused by a single attention space.

Finally, in the output layer, the model uses a feedforward neural network and regularization to achieve the final judgment of anomalies. Specifically, the output hidden state $H'$ is transformed through two layers of nonlinear transformation to obtain the prediction result:

$$Z = RELU(H'W_1 + b_1) \tag{5}$$

$$\hat{y} = \sigma(ZW_2 + b_2) \tag{6}$$

Where $W_1, W_2$ is a trainable weight, $b_1, b_2$ is a bias term, and $\sigma(\cdot)$ is a Sigmoid function that outputs anomaly probabilities in the range [0, 1]. This design enables the model to efficiently detect accounting transaction anomalies while maintaining real-time computational efficiency and leveraging deep feature representation.

## IV. EXPERIMENTAL RESULTS

### A. Dataset

In this study, the dataset selected is the Enron Email Dataset. This dataset is widely used in financial and accounting research for anomaly detection and risk analysis. It was originally derived from authentic corporate internal email communications. It contains millions of email records with timestamps, subjects, and interaction information. Because it involves many details related to transactions, communications, and business exchanges, it serves as an important public resource for the study of dynamic anomaly detection.

The dataset is characterized by its large scale and long time span. It covers communications across different departments, roles, and contexts, providing a comprehensive view of business and financial activity patterns in daily corporate operations. By modeling and analyzing these data, it is possible to identify abnormal patterns hidden within normal behavior sequences, such as atypical transaction activities or potential fraud signals. This feature makes the dataset well-suited as the experimental foundation for dynamic anomaly detection in accounting flows.

In addition, the Enron Email Dataset has been widely used in the research community to test and compare different modeling approaches. Its openness and reproducibility provide strong support for method development and performance evaluation. Although the dataset is essentially email communication data, the dynamic interaction patterns it reflects are highly consistent with the requirements of anomaly detection in corporate accounting flows. For this reason, it was chosen as the primary data source for this study.

### B. Experimental Results

This paper first gives the results of the comparative experiment, and the experimental results are shown in Table 1.

Table1. Comparative experimental results

| Model | AUC | F1-Score | Precision | Recall |
|---|---|---|---|---|
| XGBoost[47] | 0.842 | 0.781 | 0.765 | 0.798 |
| DT[48] | 0.791 | 0.732 | 0.715 | 0.749 |
| 1DCNN[49] | 0.873 | 0.812 | 0.826 | 0.799 |
| Transformer[50] | 0.902 | 0.838 | 0.846 | 0.831 |
| Ours | 0.927 | 0.864 | 0.871 | 0.857 |

From the overall results, different methods show clear differences in performance for dynamic anomaly detection in accounting transactions. The traditional decision tree method performs the weakest, with both AUC and F1-Score at low levels. This indicates that models based on a single split are unable to capture the underlying patterns of abnormal behavior when facing complex temporal features and multi-dimensional interactions. Although XGBoost improves model complexity and feature representation, it still relies on local partitioning logic within tree structures. As a result, it has limitations in both precision and recall and cannot effectively model dynamic transaction sequences on a global scale.

In comparison, deep learning methods demonstrate stronger feature learning and pattern recognition capabilities. The 1D CNN achieves results that are clearly better than traditional methods, especially showing strong performance in precision. This suggests that it has an advantage in extracting local temporal features. However, convolutional structures are inherently weak in capturing long-range dependencies and global patterns. This leads to suboptimal recall, which prevents further improvement in overall F1-Score and AUC.

The introduction of the Transformer architecture effectively alleviates this limitation. By leveraging self-attention mechanisms that capture global dependencies across the entire sequence, the Transformer surpasses earlier models in all evaluation metrics and achieves a more balanced performance. The improvement in the F1-Score is particularly noteworthy, demonstrating the model's superior ability to balance precision and recall—an aspect that is crucial in accounting transaction analysis. Low recall may result in undetected risks, while low precision could lead to excessive false alarms; achieving an optimal balance between the two is therefore vital for practical deployment. The final experimental results indicate that the proposed model attains the highest performance across all metrics. Compared with the Transformer baseline, the incorporation of structural enhancements and feature aggregation mechanisms yields consistent improvements in AUC, precision, and recall. These findings suggest that the model adapts more effectively to the complex, dynamic nature of accounting transactions and achieves more accurate anomaly detection. The performance gains are reflected not only in quantitative metrics but also in the model's demonstrated applicability to real-world financial data, providing enterprises with a dependable tool for real-time risk monitoring and prevention.

This paper gives the impact of the number of attention heads on the experimental results, and the experimental results are shown in Figure 2.

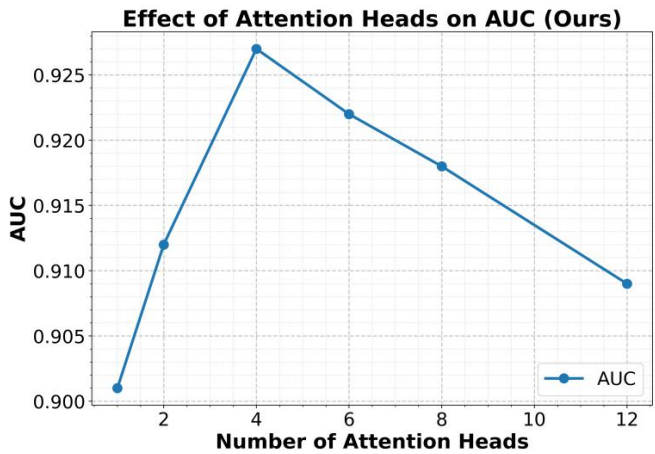

Figure 2. The impact of the number of attention heads on experimental results

From the experimental results, it can be observed that the number of attention heads has a significant impact on model performance within a certain range. When the number of heads is small, the model is limited in capturing global dependencies, which leads to weaker overall performance and lower AUC values. As the number of heads increases, the model can attend to different patterns and features of the transaction sequences in parallel from multiple subspaces. As a result, the performance reaches its peak at four attention heads, with the most noticeable improvement in AUC.

After the peak, further increasing the number of heads does not continue to improve performance. Instead, a slight decline appears. This indicates that when too many heads are used, the model may introduce redundant information or noise during feature aggregation. In this case, the decomposition of the feature space no longer provides clear benefits. Overly complex subspace partitioning not only increases computational cost but may also weaken the focus on key transaction patterns, leading to a downward trend in AUC.

Considering the context of anomaly detection in accounting transactions, a reasonable setting of attention heads is crucial for improving detection performance. The results show that in financial data, which is high-dimensional and long in sequence length, the model must strike a balance between expressive power and stability. Too few heads cannot capture complex patterns adequately, while too many heads cause feature mixing and reduce generalization ability. Therefore, a moderate configuration is required to maximize performance.

Overall, these findings reveal the sensitivity of model performance to internal hyperparameters and confirm the suitability of Transformer-based structures for dynamic accounting transaction scenarios. By carefully adjusting the number of attention heads, the model can more effectively identify potential anomalies in real time and provide accurate support for enterprise risk management. This insight offers clear reference value for future parameter optimization and practical deployment.

## V. Conclusion

This study focuses on real-time anomaly detection in accounting transactions based on Transformer architectures. It proposes an efficient framework that balances temporal modeling with global dependency modeling. In the analysis and comparison with other methods, the designed model shows clear advantages in feature representation and detection performance. By building a unified representation and adopting an efficient attention mechanism, the proposed approach not only improves detection accuracy but also demonstrates strong robustness in complex transaction environments. These findings indicate that deep temporal modeling has high applicability and value in accounting anomaly detection and provide a solid theoretical and methodological foundation for future research.

From a practical perspective, the significance of this study lies not only in structural improvements to the model but also in advancing the application of intelligent auditing and real-time risk monitoring. Traditional rule-based or single-feature statistical methods often fail when facing complex and diverse accounting data. In contrast, the proposed framework can identify potential risks within massive transaction records promptly, preventing enterprises from suffering greater losses due to delayed responses. This mechanism enhances transparency and security in financial operations and enables financial institutions to achieve stronger compliance and risk control.

In terms of broader impact, the contributions of this study extend beyond accounting transaction scenarios. The framework can be applied to other domains that involve time-series data and anomaly detection. Examples include financial transaction monitoring, insurance claim analysis, and supply chain fund flow tracking. These fields share features of large-scale data, hidden anomaly patterns, and strong real-time requirements. The proposed framework offers a feasible and efficient solution for these areas, supporting intelligent risk identification and dynamic decision-making across industries.

Looking ahead, this research provides a clear direction for future work while maintaining technical innovation. On the one hand, more lightweight structural designs and efficient optimization algorithms can be introduced to further improve computational efficiency and deployment feasibility in real-time environments. On the other hand, cross-domain transfer and multi-source data fusion can be explored, allowing the model to retain strong generalization across different enterprises and application scenarios. Ultimately, as the framework continues to improve and expand in scope, it has the potential to become a key support for intelligent financial auditing and digital corporate governance, driving industries toward a safer, more efficient, and more intelligent future.

## References

[1] A. Anandakrishnan, S. Kumar, A. Statnikov, et al., "Anomaly detection in finance: editors' introduction," *Proceedings of the KDD 2017 Workshop on Anomaly Detection in Finance*, PMLR, pp. 1-7, 2018.

[2] H. Kang and P. Kang, "Transformer-based multivariate time series anomaly detection using inter-variable attention mechanism," *Knowledge-Based Systems*, vol. 290, 111507, 2024.

[3] A. K. T. Sharmal and A. Daha, "Big data analytics anomaly transformer in financial fraud detection in BankSim datasets," 2025.

[4] L. Lian, "Automatic elastic scaling in distributed microservice environments via deep Q-learning," Transactions on Computational and Scientific Methods, vol. 4, no. 4, 2024.

[5] W. Zhu, "Adaptive container migration in cloud-native systems via deep Q-learning optimization," Journal of Computer Technology and Software, vol. 3, no. 5, 2024.

[6] B. Fang and D. Gao, "Domain-adversarial transfer learning for fault root cause identification in cloud computing systems," arXiv preprint arXiv:2507.02233, 2025.

[7] Y. Sun, R. Meng, R. Zhang, Q. Wu and H. Wang, "A deep Q-network approach to intelligent cache management in dynamic backend environments," 2025.

[8] N. Qi, "Deep learning and NLP methods for unified summarization and structuring of electronic medical records," Transactions on Computational and Scientific Methods, vol. 4, no. 3, 2024.

[9] X. Yan, Y. Jiang, W. Liu, D. Yi and J. Wei, "Transforming Multidimensional Time Series into Interpretable Event Sequences for Advanced Data Mining," 2024 5th International Conference on Intelligent Computing and Human-Computer Interaction (ICHCI), pp. 126-130, 2024.

[10] Y. Zou, N. Qi, Y. Deng, Z. Xue, M. Gong and W. Zhang, "Autonomous resource management in microservice systems via reinforcement learning," arXiv preprint arXiv:2507.12879, 2025.

[11] Y. Li, S. Han, S. Wang, M. Wang and R. Meng, "Collaborative evolution of intelligent agents in large-scale microservice systems," arXiv preprint arXiv:2508.20508, 2025.

[12] Y. Zhang and B. Duan, "Accounting data anomaly detection and prediction based on self-supervised learning," Frontiers in Applied Mathematics and Statistics, vol. 11, 1628652, 2025.

[13] W. Cui, "Vision-oriented multi-object tracking via transformer-based temporal and attention modeling," Transactions on Computational and Scientific Methods, vol. 4, no. 11, 2024.

[14] D. Gao, "High fidelity text to image generation with contrastive alignment and structural guidance," arXiv preprint arXiv:2508.10280, 2025.

[15] C. Liu, Q. Wang, L. Song and X. Hu, "Causal-aware time series regression for IoT energy consumption using structured attention and LSTM networks," 2025.

[16] Y. Xing, "Bootstrapped structural prompting for analogical reasoning in pretrained language models," , 2024.

[17] Z. Xue, "Dynamic structured gating for parameter-efficient alignment of large pretrained models," Transactions on Computational and Scientific Methods, vol. 4, no. 3, 2024.

[18] D. Wu and S. Pan, "Joint modeling of intelligent retrieval-augmented generation in LLM-based knowledge fusion," 2025.

[19] W. Zhu, "Fast adaptation pipeline for LLMs through structured gradient approximation," Journal of Computer Technology and Software, vol. 3, no. 6, 2024.

[20] X. Quan, "Structured path guidance for logical coherence in large language model generation," , 2024.

[21] R. Zhang, "Privacy-oriented text generation in LLMs via selective fine-tuning and semantic attention masks," , 2025.

[22] C. Hu, Z. Cheng, D. Wu, Y. Wang, F. Liu and Z. Qiu, "Structural generalization for microservice routing using graph neural networks," arXiv preprint arXiv:2510.15210, 2025.

[23] Y. Zou, "Federated distillation with structural perturbation for robust fine-tuning of LLMs," , 2025.

[24] H. Hemati, M. Schreyer and D. Borth, "Continual learning for unsupervised anomaly detection in continuous auditing of financial accounting data," arXiv preprint arXiv:2112.13215, 2021.

[25] X. Hu, Y. Kang, G. Yao, T. Kang, M. Wang and H. Liu, "Dynamic prompt fusion for multi-task and cross-domain adaptation in LLMs," arXiv preprint arXiv:2509.18113, 2025.

[26] X. Song, Y. Liu, Y. Luan, J. Guo and X. Guo, "Controllable abstraction in summary generation for large language models via prompt engineering," arXiv preprint arXiv:2510.15436, 2025.

[27] T. Su, R. Li, B. Liu, et al., "Anomaly detection and risk early warning system for financial time series based on the WaveLST-Trans model," 2025.

[28] P. Dubey, P. Dubey and P. N. Bokoro, "A unified transformer–BDI architecture for financial fraud detection: Distributed knowledge transfer across diverse datasets," Forecasting, vol. 7, no. 2, pp. 31, 2025.

[29] Y. Wang, "Entity-aware graph neural modeling for structured information extraction in the financial domain," Transactions on Computational and Scientific Methods, vol. 4, no. 9, 2024.

[30] Z. Liu and Z. Zhang, "Graph-based discovery of implicit corporate relationships using heterogeneous network learning," Journal of Computer Technology and Software, vol. 3, no. 7, 2024.

[31] Q. Sha, T. Tang, X. Du, J. Liu, Y. Wang and Y. Sheng, "Detecting credit card fraud via heterogeneous graph neural networks with graph attention," arXiv preprint arXiv:2504.08183, 2025.

[32] W. Xu, M. Jiang, S. Long, Y. Lin, K. Ma and Z. Xu, "Graph neural network and temporal sequence integration for AI-powered financial compliance detection," 2025.

[33] Y. Zi, M. Gong, Z. Xue, Y. Zou, N. Qi and Y. Deng, "Graph neural network and transformer integration for unsupervised system anomaly discovery," arXiv preprint arXiv:2508.09401, 2025.

[34] Q. R. Xu, W. Xu, X. Su, K. Ma, W. Sun and Y. Qin, "Enhancing systemic risk forecasting with deep attention models in financial time series," 2025.

[35] Z. Xu, X. Liu, Q. Xu, X. Su, X. Guo and Y. Wang, "Time series forecasting with attention-augmented recurrent networks: A financial market application," 2025.

[36] Q. Sha, "Hybrid deep learning for financial volatility forecasting: An LSTM-CNN-transformer model," Transactions on Computational and Scientific Methods, vol. 4, no. 11, 2024.

[37] M. Gong, "Modeling microservice access patterns with multi-head attention and service semantics," Journal of Computer Technology and Software, vol. 4, no. 6, 2025.

[38] Y. Yao, Z. Xu, Y. Liu, K. Ma, Y. Lin and M. Jiang, "Integrating feature attention and temporal modeling for collaborative financial risk assessment," arXiv preprint arXiv:2508.09399, 2025.

[39] J. Hu, B. Zhang, T. Xu, H. Yang and M. Gao, "Structure-aware temporal modeling for chronic disease progression prediction," arXiv preprint arXiv:2508.14942, 2025.

[40] L. Dai, "Contrastive learning framework for multimodal knowledge graph construction and data-analytical reasoning," Journal of Computer Technology and Software, vol. 3, no. 4, 2024.

[41] W. Cui, "Unsupervised contrastive learning for anomaly detection in heterogeneous backend system," Transactions on Computational and Scientific Methods, vol. 4, no. 7, 2024.

[42] Y. Wang, H. Liu, N. Long and G. Yao, "Federated anomaly detection for multi-tenant cloud platforms with personalized modeling," arXiv preprint arXiv:2508.10255, 2025.

[43] H. Wang, "Causal discriminative modeling for robust cloud service fault detection," Journal of Computer Technology and Software, vol. 3, no. 7, 2024.

[44] Z. Liu and Z. Zhang, "Modeling audit workflow dynamics with deep Q-learning for intelligent decision-making," Transactions on Computational and Scientific Methods, vol. 4, no. 12, 2024.

[45] M. Jiang, S. Liu, W. Xu, S. Long, Y. Yi and Y. Lin, "Function-driven knowledge-enhanced neural modeling for intelligent financial risk identification," 2025.

[46] X. Zhang and Q. Wang, "EEG anomaly detection using temporal graph attention for clinical applications," Journal of Computer Technology and Software, vol. 4, no. 7, 2025.

[47] S. Kabane, "Impact of sampling techniques and data leakage on XGBoost performance in credit card fraud detection," arXiv preprint arXiv:2412.07437, 2024.

[48] E. Flondor, L. Donath and M. Neamtu, "Automatic card fraud detection based on decision tree algorithm," *Applied Artificial Intelligence*, vol. 38, no. 1, 2385249, 2024.

[49] M. T. R. Mazumder, M. S. H. Shourov, I. Rasul, et al., "Anomaly detection in financial transactions using convolutional neural networks," *Journal of Economics, Finance and Accounting Studies*, vol. 7, no. 2, pp. 195-207, 2025.

[50] D. Yadav, S. Zhang and T. Jin, "Transformer based anomaly detection on multivariate time series subledger data," 2023.